\documentclass{article}

\usepackage{microtype}
\usepackage{graphicx}
\usepackage{subfigure}
\usepackage{booktabs} 

\usepackage{hyperref}

\usepackage[accepted]{icml2018_arxiv}
\usepackage{amsmath}
\usepackage{graphicx, caption, subfigure} 
\usepackage{amsfonts}
\usepackage{multicol}

\icmltitlerunning{TensOrMachine}
\graphicspath{{./figs/}}

\newcommand{\tr}[1]{\textcolor{black}{#1}}

\begin{document}

\twocolumn[
\icmltitle{TensOrMachine: Probabilistic Boolean Tensor Decomposition}



\icmlsetsymbol{equal}{*}
\begin{icmlauthorlist}
\icmlauthor{Tammo Rukat}{stats,ati}
\icmlauthor{Chris C.~Holmes}{stats,ati}
\icmlauthor{Christopher Yau}{ati,bmh}
\end{icmlauthorlist}

\icmlaffiliation{stats}{Department of Statistics, University of Oxford, UK}
\icmlaffiliation{ati}{The Alan Turing Institute, London, UK}
\icmlaffiliation{bmh}{Centre for Computational Biology, Institute of Cancer and Genomic Sciences, University of Birmingham, UK} 

\icmlcorrespondingauthor{Tammo Rukat}{tammo.rukat@stats.ox.ac.uk}

\icmlkeywords{Matrix factorisation, Bayesian modelling, Genomics, Binary data}

\vskip 0.3in
]



\printAffiliationsAndNotice{}  

\begin{abstract} 
\setcounter{footnote}{1}
Boolean tensor decomposition approximates data of multi-way binary relationships
as product of interpretable low-rank binary factors, following the rules of Boolean algebra.
Here, we present its first probabilistic treatment.
We facilitate scalable sampling-based posterior inference by exploitation
of the combinatorial structure of the factor conditionals.
Maximum a posteriori decompositions feature higher accuracies than existing techniques throughout a wide range of simulated conditions.
Moreover, the probabilistic approach facilitates the treatment of missing data and enables model selection with much greater accuracy. We investigate three real-world data-sets. First, temporal interaction networks in a hospital ward and behavioural data of university students demonstrate the inference of instructive latent patterns. Next, we decompose a tensor with more than 10 billion data points, indicating relations of gene expression in cancer patients. Not only does this demonstrate scalability, it also provides an entirely novel perspective on relational properties of continuous data and, in the present example, on the molecular heterogeneity of cancer. Our implementation is available on GitHub\footnote{\tiny \url{https://github.com/TammoR/LogicalFactorisationMachines}}.

\end{abstract}

\section{Introduction}

Matrix decomposition methods such as factor analysis are a widely used class of models for unsupervised learning. Typically, they operate on two-way matrices with rows thought of as objects and columns thought of as properties. The decomposition amounts to computing two low-rank matrices, one that contains prototypes of properties (loadings) and one that denotes a compressed representation of each object as a combination of the prototypes (factors). 
However, these methods are ill-suited for data with higher arity such as ternary interactions. Examples include network interactions at different time-points, gene-gene interactions for different individuals or any two-way data under different experimental conditions.
Such data requires methods that specifically account for the higher-order relationship and that are commonly referred to as \textit{tensor decomposition}.

Boolean tensor decomposition factorises a $K$-way binary tensor $X \in \{0,1\}^{N_1 \times \ldots \times N_K}$, into $K$ binary factor matrices $F_k \in \{0,1\}^{N_k \times L}$, using the Boolean algebra such that
\begin{align}
  x_{[n]} \approx \bigvee \limits_{l=1}^L \left[ 
  	\bigwedge\limits_{n \in [n]} f_{nl} \right]\;. \label{eq:bool}
\end{align}
Here, $x_{[n]}$ is a single tensor entry and $[n]$ denotes a tuple of indices $[n_1, \ldots, n_K]$. We call $k=1\ldots K$ the \textit{modes} of the tensor, $\vee$ and $\wedge$ denote the logical \texttt{OR} and the logical \texttt{AND} operation, respectively. 
\begin{figure}	
	\includegraphics[width=\linewidth]{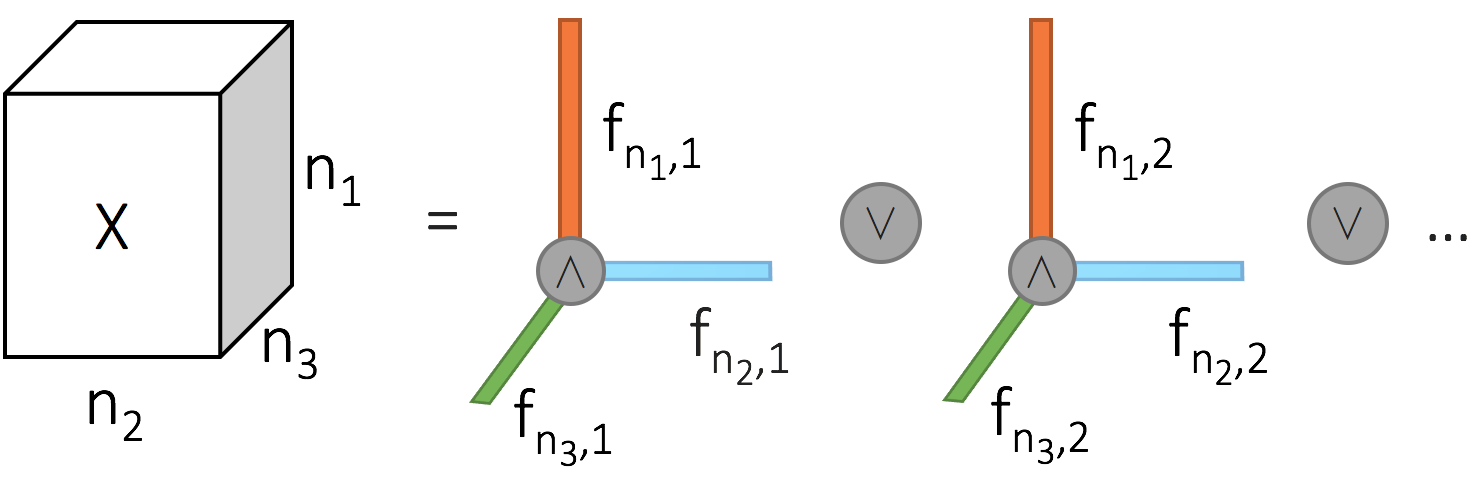} 
	\caption{Graphical intuition for 3-way Boolean tensor decomposition. \tr{Logical conjunction ($\wedge$) of Boolean vectors (columns of factor matrices) generates Rank-1 tensors. The full tensor is the logical disjunction ($\vee$) of rank-1 tensors.}
	\label{fig:toy_example}}
\end{figure}
In plain English, eq.\ \eqref{eq:bool} says that an observation is $1$, if and only if there exist one or more latent dimensions in which all corresponding factors are $1$. This leads to easily interpretable factor matrices, $F_k$, where each latent dimension denotes a subset of entries along mode $k$, for instance subsets of objects, properties and conditions that occur jointly in the data.
\tr{This can be thought of as a particular type of canonical polyadic (CP) decomposition, as
illustrated in Fig.\ \ref{fig:toy_example}.
A different graphical intuition starts from the factor rows and considers the 3-way Boolean tensor product as a three-stage template assignment procedure. Rows of a factor matrix $F_1$ are one-dimensional binary templates of size of the first tensor mode.} Rows of the second factor matrix $F_2$ indicates possible patterns of appearance of these templates along the second tensor dimensions. In the same manner, the third factor matrix, $F_3$, indicates which disjunction of these 2D patterns appears in each slice of the tensor, and so forth. 

We present the first probabilistic approach to Boolean tensor decomposition, the \textit{TensOrMachine}, featuring distinctly improved accuracy compared to the previous state-of-the-art methods. We develop scalable, sampling-based posterior inference and infer distributions over factor matrices which enables full quantification of uncertainty. In contrast to previous approaches, the probabilistic framework readily treats missing data, allows for tensor completion and integration of prior information.
Importantly, the latent representations are amenable to informative and intuitive interpretation.

\section{Related Work}
Tensor decomposition methods are widely used in many domains of application as reviewed by~\citet{Kolda2009}.
Boolean Tensor decomposition was first introduced in the Psychometric literature by \citet{Leenen1999} and
has received lasting attention, following the formal study by \citet{Miettinen2011}. The author shows that computing the optimal decomposition is NP-hard and proposes an alternating least squares heuristics as approximate strategy. A different approach based on a random-walk procedure is described by~\citet{Erdos2013a} and trades accuracy in the factors for computational scalability. More recently a distributed Apache Spark 
implementation of the alternating least squares approach has been brought forward~\cite{Park2017}. It demonstrates that alternating leasts squares is the state-of-the-art for computing accurate decompositions and serves as baseline method for our experiments.
Boolean Matrix Factorisation~\cite{miettinen2006_discr-basis-probl} shares the logical structure of Boolean tensor decomposition. It has many real-world applications such as collaborative filtering~\cite{su2009_survey-collab} and computer vision \cite{lazaro-gredilla2016_hierar} and has previously been approached under a probabilistic perspective~\cite{ravanbakhsh2015_boolean-matrix,Rukat2017}. Despite the sustained theoretical interest, there have been only few practical applications of Boolean tensor decomposition, as for instance for information extraction~\cite{Erdos2013} and clustering \cite{Metzler2015}. One of the contributions of this work is the presentation of Boolean Tensor decomposition as an interpretable, versatile and scalable analysis method.






\section{The TensOrMachine}

\subsection{Model Description}

We propose a probabilistic generative process for the model described in eq.\ \eqref{eq:bool}. Each tensor entry, $x_{[n]} = x_{[n_1,\ldots,n_K]}$, is a Bernoulli random variable that equals 1 with a probability greater $\frac{1}{2}$ if the following holds true
\begin{align}
	 \exists \; l: f_{nl} = 1\; \forall\; n \in [n_1,\ldots,n_K].\label{eq:condition}
\end{align}
The logistic sigmoid $\sigma(x)=(1+e^{-x})^{-1}$ has the convenient property $\sigma(-x)=1-\sigma(x)$ and lets us readily parametrise the corresponding likelihood
\begin{align}
	p(x_{[n]}|.) = \sigma\left[\lambda \tilde{x}_{[n]} \left( 1 - 2 
		\prod\limits_l \left[1-\prod\limits_{n \in [n]} f_{nl} \right] \right) \right].\label{eq:likelihood}
\end{align}
The term inside parenthesis evaluates to 1 if the condition in eq.~\eqref{eq:condition} is met and to -1, otherwise. Throughout this work, we use a tilde to denote the mapping from $\{0,1\}$ to $\{-1,1\}$ such that $\tilde{x}=2x-1$. The noise is parametrised by $\lambda \in \mathbb{R}^+$, such that for $\lambda \rightarrow 0$ the likelihood is constant, independent of the factors and for $\lambda \rightarrow \infty$ all probability mass is put on the deterministic Boolean tensor product following eq.~\eqref{eq:bool}. 
We can specify Bernoulli priors on the observations $x_{[n]}$ or choose more structured prior distributions. A beta-prior on $\sigma(\lambda)$ is a computationally convenient choice as we discuss in the following section. However, in the relevant regime of thousand and more data-points such priors are easily outweighed by the observed data. We therefore assume constant prior distributions in the following derivations and experiments.

\subsection{\tr{Relation to Noisy-OR Models}}
Boolean decomposition models have an interesting relation to the noisy-OR, a canonical model for multi-causal interactions, which makes the assumption of independence between the inhibition of latent causes~\cite{Pearl1988}. TensOrM violates this assumption with inhibition occurring on the event-level rather than on the cause-level. 
As an example, in analogy to the data presented in Section.~\ref{subsec:students}, consider the event of a students attending a lecture. We assume that this has two possible causes. They may (i) use the opportunity to meet fellow students (ii) have a particular question that they wish to clarify in class. An inhibition on the level of causes may occur if they found a satisfying answer to their question in a textbook. Then, the desire to meet their friends still contributes to the likelihood of attending class. This naturally modelled by the noisy-OR. In contrast, inhibition on the event-level may occur if a traffic disruption makes it impossible to get to campus. This is naturally described by a Boolean factor model, where any additional causes to attend the lecture will not contribute to the likelihood. 
We argue that for many practical purposes the latter is an interesting alternative.
Even if we were to believe in the independent inhibition of latent causes, usually only one latent cause triggers an event. 
Compared to the noisy-OR, TensOrM acts as an Occam's razor and enforces sparse and simple explanations, \tr{aiming to associate exactly one hidden cause with every event}.


\subsection{Posterior Inference}\label{sec:inference}

\tr{We condition any entry of a factor matrix, $f_{n_kl}$, on the state of all other parameters which yields the full conditional probability for Gibbs sampling}:
\begin{align}
	p(f_{n_kl}|.) =
	\sigma\Bigg(\lambda \tilde{f}_{n_kl}
		\sum\limits_{\substack{[n] \\ n_k \text{fixed}}} \tilde{x}_{[n]}
		M_{(n_k,l) \rightarrow [n]} \Bigg)\;. \label{eq:conditional}
\end{align}
We give a formal derivation in the Supplementary Information A and provide an intuitive explanation in the following.
The sigmoid maps from the real line to the interval of $[0,1]$. The sum in its argument is taken over all observations $x_{[n]}$ whose likelihood may depend on $f_{n_kl}$. These observations are given by the $(K{-}1)$--way sub-tensor of X with dimension $n_k$ fixed and correspond to all children of $f_{n_kl}$ in a graphical model representation. For each of them, the term inside the sigmoid contributes values in $\{-\lambda, 0, \lambda\}$. This depends on whether or not $\tilde{f}$ and $\tilde{x}$ are aligned and an another indicator variable $M_{(n_k,l) \rightarrow [n]}$ that take values in $\{0,1\}$ and denotes whether the state of $f_{n_kl}$ has any relevance for the likelihood of $x_{[n]}$. It takes the form
\begin{align}
	M_{(n_k,l) \rightarrow [n]} = \bigg(\prod\limits_{n \in [n]/n_k} f_{nl}\bigg)
		\prod\limits_{l'\neq l} \bigg(1{-}\prod\limits_{n \in [n]}f_{nl'}\bigg).\label{eq:M}
\end{align}
This variable is again composed of two indicators.
The \textit{first term} is a product over the state of all co-parents of $f_{n_kl}$ to observation $x_{[n]}$ in the same latent dimension $l$. Following the rules of the Boolean product, all these co-parents need to be one in order for $f_{n_kl}$ to contribute to the likelihood of $x_{[n]}$. 
This corresponds to the \texttt{AND}-operation in eq.~\eqref{eq:bool} that evaluates to zero if any its arguments are zero. The \textit{second term} is a product of the co-parents in all other latent dimensions and evaluates to zero if any of them \textit{explains away} observation $x_{[n]}$. Following the rules of the Boolean product, a single latent dimension is sufficient to explain an observation and makes its likelihood independent of the state of $f_{n_kl}$. This corresponds to the \texttt{OR}-operation in eq.~\eqref{eq:bool} that evaluates to one if any of its arguments is one. 
It is crucial for the speed of our implementation that eq.\ \eqref{eq:M}, and thus eq.\ \eqref{eq:conditional}, can be rapidly evaluated. In particular, discovering a zero in any of the two terms of eq.\ \eqref{eq:M} suffices for the whole expression to be zero and avoids the necessity of considering the remainder of the Markov blanket.
Pseudocode for the computational procedure is given in Algorithm~\ref{alg:update_f}. Note that updates of $f_{n_kl}$ can trivially be computed in parallel across $n_k$ for a fixed tensor-mode $k$. \tr{The computation time scales linearly in each tensor dimension and sub-linear in the latent dimension.}
\begin{algorithm}
  \caption{Computation of the full conditional of $f_{n_kl}$}
  \label{alg:update_f}
\begin{algorithmic}[tb]
  \STATE{$\text{m}=0$ // \textit{initialise integer count for the sum in eq.\ \eqref{eq:conditional}}}
  \FOR{[n] in all tensor indices with slice $n_k$ fixed}
  \STATE // \textit{check relevance of $f_{n_kl}$ for $x_{[n]}$.}
  \FOR{n in [n]}
  \IF{$f_{nl} = 0$}
  \STATE // \textit{$f_{n_kl}$ has no relevance for $x_{[n]}$.}
  \STATE continue with next [n]
  \ENDIF
  \ENDFOR
  \STATE // \textit{check for explaining away.}
  \FOR{$l'\;\text{in}\;1,\ldots, L \text{ except } l$}
    \FOR{n in [n]}
      \IF{$f_{nl'}=0$}
        \STATE continue with next l'
      \ENDIF
    \STATE // $x_{[n]}$ \textit{is explained away.}
    \STATE continue (next [n])
  \ENDFOR
  \ENDFOR
  \STATE $\text{m} =  \text{m} + \tilde{x}_{nd} $
  \ENDFOR \STATE{$p(f_{n_kl}|.) = \left(1+\exp\left(-\lambda\cdot\tilde{f}_{n_kl}\cdot\text{m} \right)\right)^{-1}$}
\end{algorithmic}
\end{algorithm}

After a sampling all factor entries from their full conditional, we set the noise-parameter, $\lambda$, to its conditional MAP estimate and repeat until convergence before drawing posterior samples. The conditional MLE of $\lambda$ is available in closed form and given by the logit of the fraction of data-points that are correctly reconstructed by the deterministic Boolean product of the current state of the factors. As a prior for $\sigma(\lambda)$, it is natural to employ a Beta-distribution, $p(\sigma(\lambda))=\text{Beta}(\sigma(\lambda)|\alpha,\beta)$. This intuitively affects the MAP estimate, adding $\alpha$ correctly predicted data points and $\beta$ incorrectly predicted data points, such that
\begin{equation}
\sigma(\lambda)_{\text{MAP}} = \frac{\alpha + \sum_{[n]} \mathcal{I}\Big(x_{[n]} = \bigvee_l \Big[ 
    \bigwedge\limits_{n \in [n]} f_{nl} \Big]\Big)}{
  \alpha +\beta + \prod\limits_{k=1}^{K} N_k}\;.\label{eq:lbda_map}
\end{equation}
Here, the indicator $\mathcal{I}$ evaluates to 1 if its arguments is true and to 0 otherwise.
We see that a uniform prior, $p(\sigma(\lambda))=\text{Beta}(\alpha{=}1,\beta{=}1)$, corresponds to applying Laplace's rule of succession to the maximum likelihood estimate.

\subsection{Tensor Reconstruction Methods and Missing Data}\label{subsec:recon}
An important application of latent variable models is the prediction of missing observations, e.g.\ in collaborative filtering or data imputation problems~\cite{su2009_survey-collab}.
Our probabilistic construction allows us to deal with missing data in a unified way.
When computing updates for the latent factors, we can marginalise over any missing observations. Practically, this is equivalent to setting any missing entries $\tilde{x}_{[n]}$ to $0$, such that they do not contribute to the conditional probability in eq.~\eqref{eq:conditional} and contribute a factor of $\frac{1}{2}$ to the likelihood in eq.~\eqref{eq:likelihood}. Thus, encoding observed data as $\{-1,1\}$ and unobserved data as 0 is the most convenient choice for any practical implementation. For MAP updates of $\lambda$, following eq.\ \eqref{eq:lbda_map}, unobserved data-points points are simply excluded.

For a direct comparison of predictions to previous methods that only provide a point estimate of the factors, we can reconstruct the data based on the Boolean product of the factor MAP estimates. We use the \tr{marginal MAP of each factor entry, $f_{n_kl} \in [0,1]$ and find no relevant difference to using the joint MAP.} If instead we want to use the full posterior, we can determine the reconstruction of $X$ by rounding the posterior predictive,
\begin{equation}
  \frac{1}{S} \sum\limits_{s=1}^M p(x_{[n]}| F_1^{(s)}, \ldots, F_K^{(s)})\;,\label{eq:predictive}
\end{equation}
to the closest binary value. Here, $(F_1^{(s)}, \ldots, F_K^{(s)})$ is a posterior sample. Yet another alternative is to compute the estimate from the factor matrix mean, thus taking posterior uncertainty but no higher-order correlations into account. Denoting the posterior mean of a factor entry as $\hat{f}$, we have
$p(x_{[n]}=1|.) \approx 1 - \prod_l (1 - \prod_{n\in [n]} \hat{f}_{nl})$.
The predictive accuracy of this computationally cheap approximation is on par with the more expensive posterior predictive estimate as we show empirically in the next section and in Fig.\ \ref{subfig:comparsion_recon}.

\section{Experiments on Simulated Data}
\subsection{Random Tensor Decomposition}

\begin{figure*}
\begin{minipage}[c]{0.49\textwidth}
\includegraphics[width=.95\linewidth]{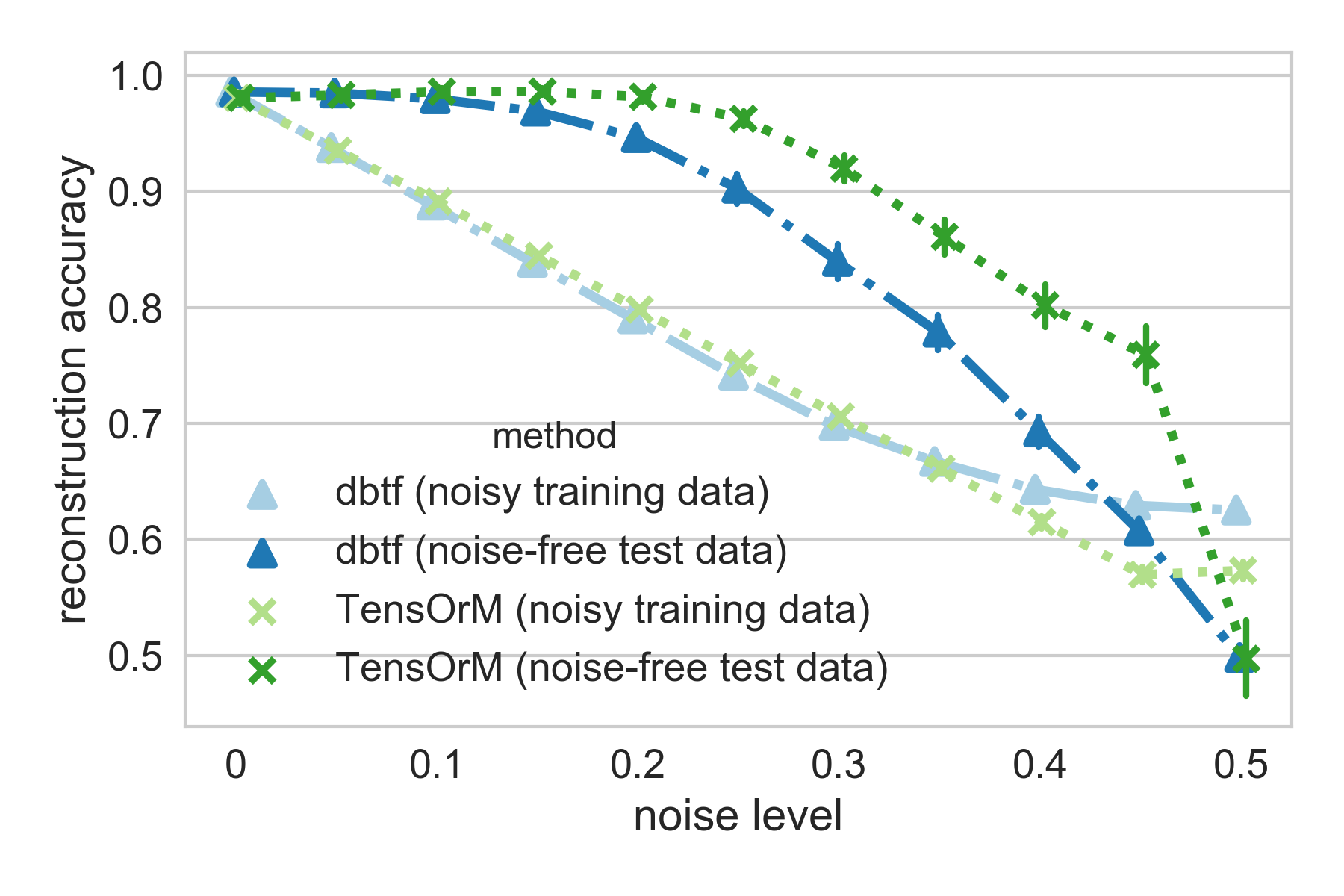}
\includegraphics[width=.95\linewidth]{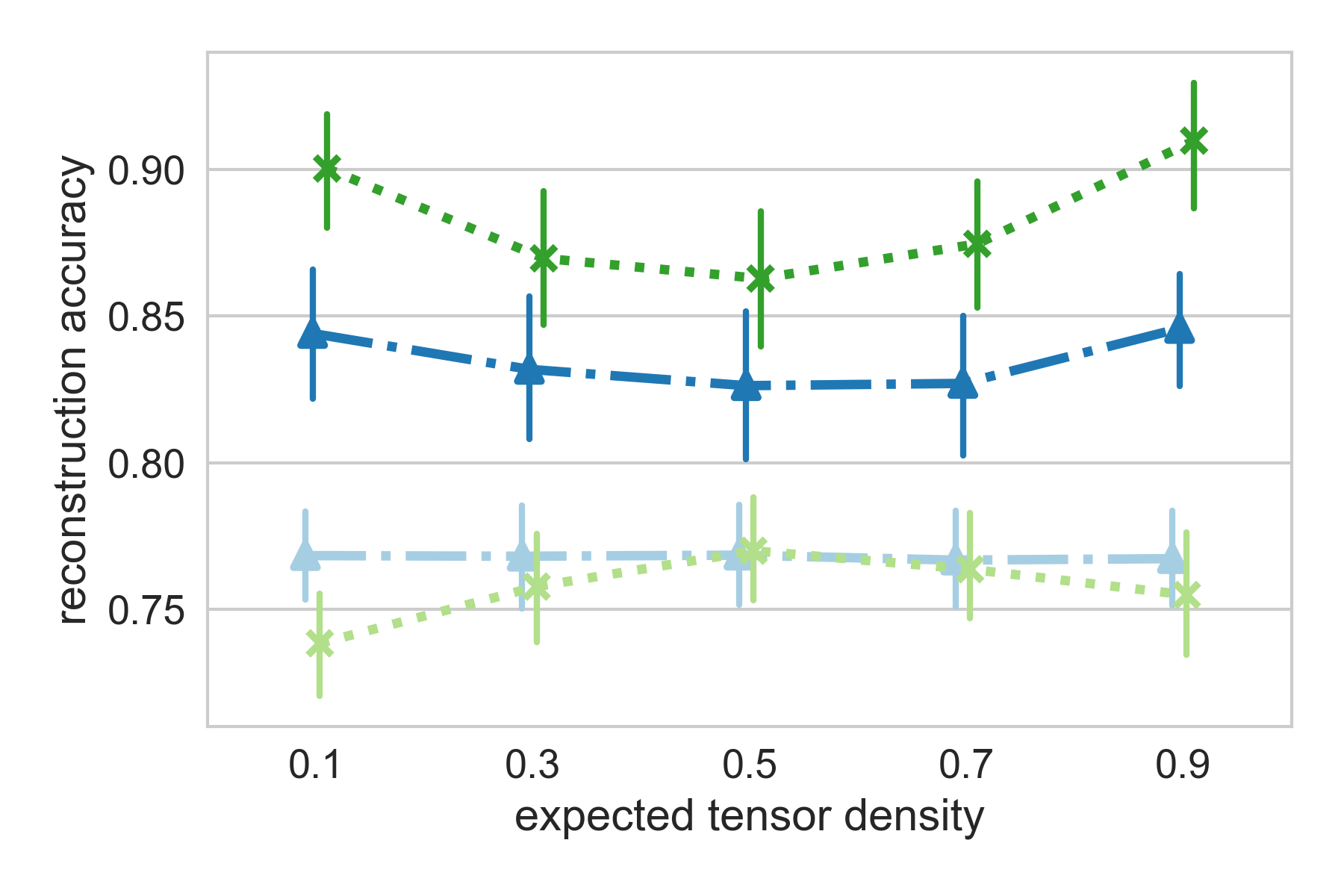}
\subfigure[\textbf{Random tensor reconstruction accuracy} under variation of the noise level (top), the expected tensor density (centre) and the underlying tensor rank (bottom). Averages are taken across all shown combinations of the other two parameters and across ten random tensors for each such configuration. Test accuracy on noise-free data is shown in solid colours, training accuracy on noisy data in faint colours. Compare to Fig.\ 8 by \citet{Park2017}.\label{fig:benchmarks}]{\includegraphics[width=.95\linewidth]{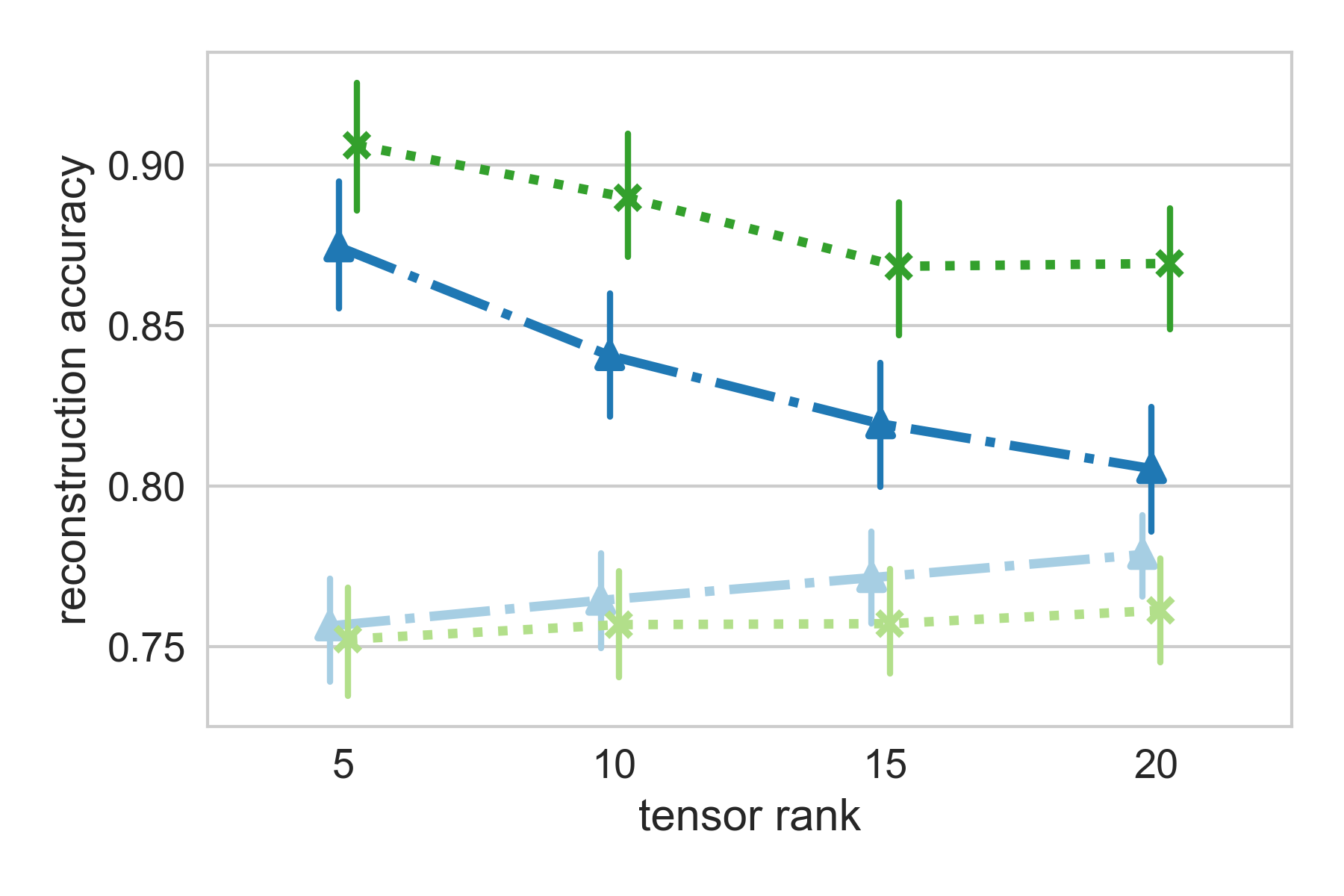}}
\end{minipage}
\hfill
\begin{minipage}[c]{0.49\textwidth}
\subfigure[\textbf{Comparison of reconstruction methods} -- We compare predictions based on TensOrM posterior predictive, the factor matrix MAP, and mean as described in Section~\ref{subsec:recon} to the the dbtf baseline. M1 indicates the performance margin that is due to finding more accurate factor matrices, M2 denotes the margin that is due to posterior averaging. The underlying tensors are of rank 10. The results are averaged across tensor densities and 10 random repetitions.]{\label{subfig:comparsion_recon}\includegraphics[width=\linewidth]{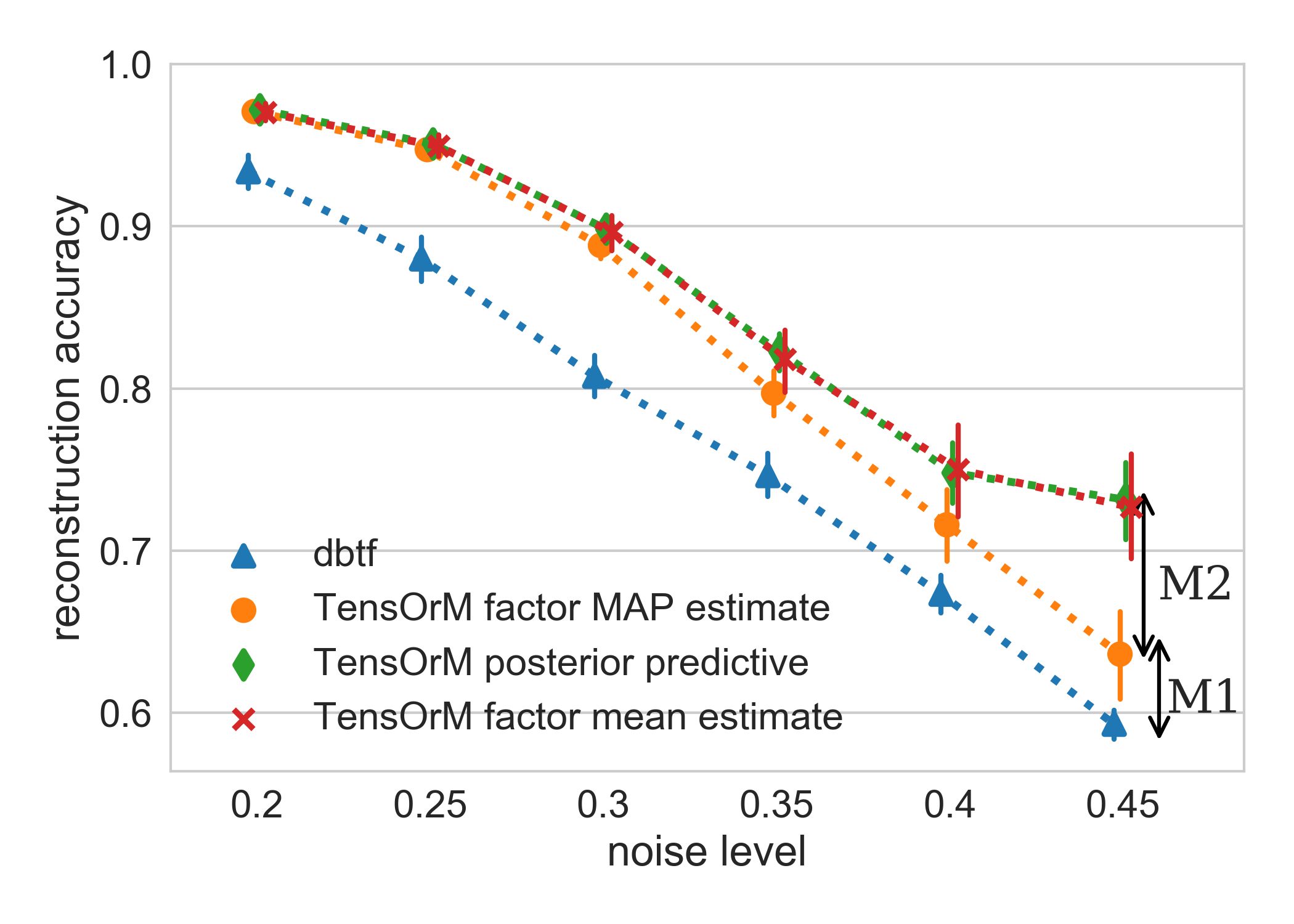}}
\subfigure[\textbf{Robustness of training performance under random perturbations} -- We show reconstruction accuracy on the \textit{noisy training data} based on the factor matrix MAP estimate under random flips of the factor entries. The underlying tensor has a noise level of $40\%$ and a expected density of 10\%. We perform 10 random repetitions but standard deviations are too small to be visible. TensOrM recovers exactly the expected reconstruction accuracy of 60\%, while dbtf overfits to the training data. Its reconstruction accuracy deteriorates rapidly under random perturbations, indicating that it has converged to a comparatively narrow mode of the posterior.]{\label{subfig:perturbation}\includegraphics[width=\linewidth]{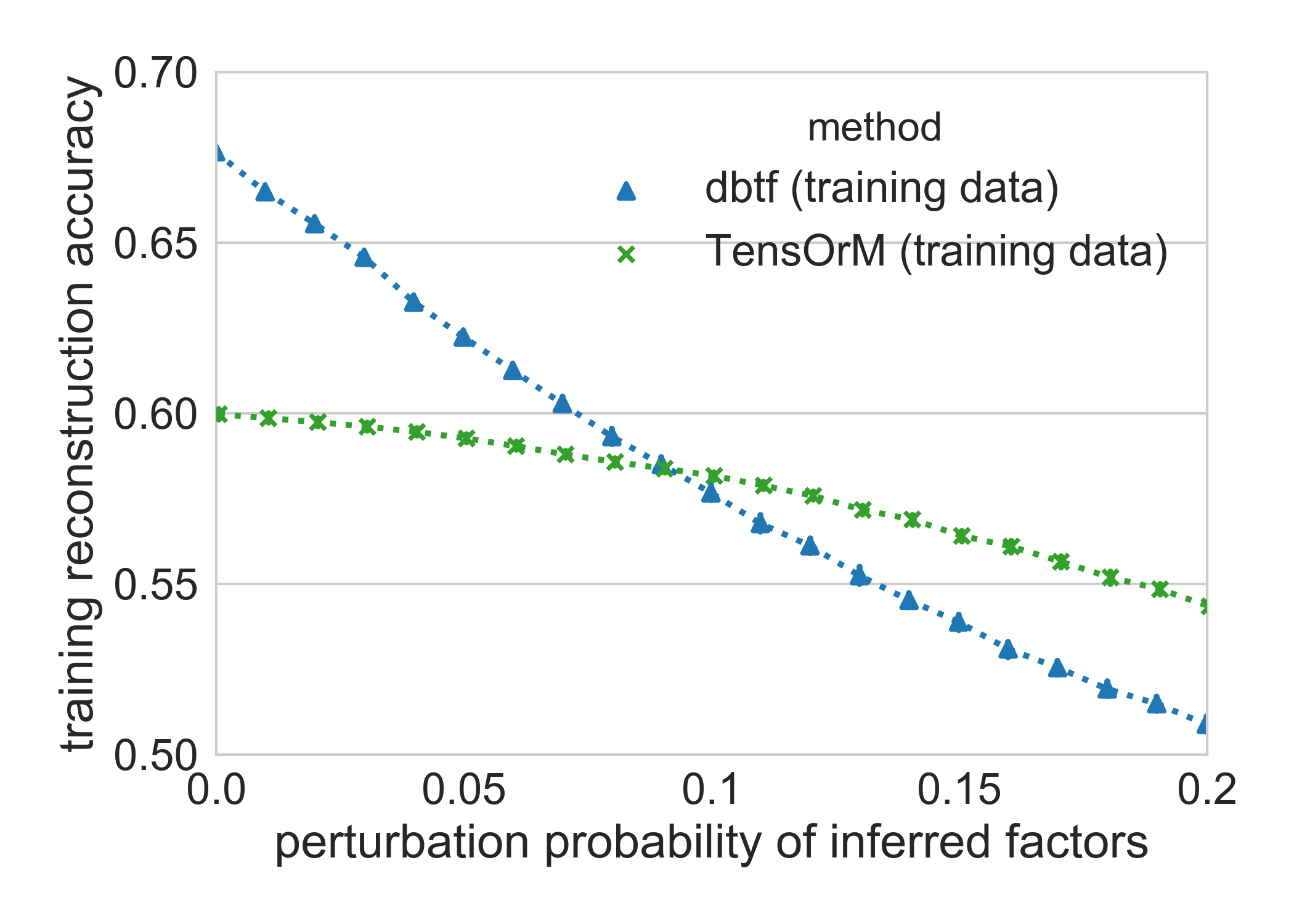}}
\end{minipage}
	\caption{
	The TensOrMachine is compared to the previous state-of-the-art method, \textit{distributed Boolean tensor decompositions (dbtf)}, and properties of its solutions are investigated. All examples are based on $20{\times}20{\times}20$ tensors generated from the Boolean product of i.i.d Bernoulli random matrices. Noise is introduced by flipping every entry in the tensor with a given probability. The reconstruction accuracy indicates the fraction of correctly reconstructed entries in the noise-free test data or in the noisy training data. Reconstructions are based on rounding the posterior predictive to the closest value if not mentioned otherwise. All plots indicate means and standard deviations (some are too small to be visible).
	\label{fig:properties_comparison}}
\end{figure*}
\begin{figure*}[!t]
	\centering
	\includegraphics[width=\linewidth]{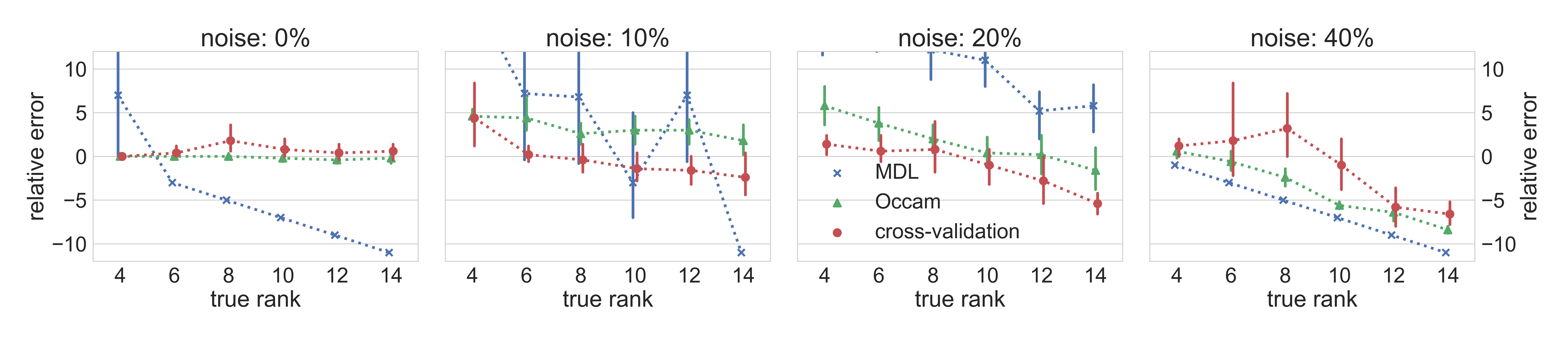}
	\caption{\textbf{Model selection accuracy} in terms of the relative error in predicting the tensor rank. Model selection with an Bayesian Occam's razor or cross validation (Section~\ref{subsec:model_selection}) outperform the previously suggested MDL approach.
	\label{fig:model_selection}}
\end{figure*}


We first demonstrate the capabilities of TensOrM on simulated data and compare to the state-of-the-art method \textit{distributed boolean tensor factorisation} (dbtf)~\cite{Park2017}. We simulate random 3-way tensors, $X$, of size $20{\times}20{\times}20$ and vary rank $L$, expected density $E(X)$ and noise-level.
To this end, we take the Boolean product between binary i.i.d.\ random matrices, $F_k$, of size $20{\times}L$, such that the expected tensor density is given by $ E(X) = 1 - \left[ 1-(E(F_k))^3 \right]^L $. We introduce noise by flipping each entry in the tensor with a given probability. 
Posterior samples of the factors are drawn, following the procedure described in Section~\ref{sec:inference} \tr{with $\lambda$ initialised to $0.5$ and the initial factors drawn i.i.d. Bern($0.5$)}. The reconstruction of $X$ is determined based on the posterior predictive following eq.\ \eqref{eq:predictive}.
The reconstruction accuracy is computed as the fraction of correctly reconstructed entries in the noise-free tensor and is shown across a variety of conditions in Fig.\ \ref{fig:benchmarks}.
Our method achieves distinctly higher accuracies throughout all conditions. The margin becomes bigger for very noisy data, as well as for higher tensor ranks and for particularly dense or particularly sparse data.

Can the superior reconstruction performance be explained by implicit model averaging when computing the posterior predictive following eq.~\eqref{eq:predictive}?
In order to address this conjecture, Fig.\ \ref{subfig:comparsion_recon} compares the reconstruction accuracies of the posterior predictive compared to the reconstruction based solely on the MAP estimates of each factor.  We can see, that posterior averaging plays an important role only in scenarios with high noise levels. The main performance gain, however, is simply due to a more accurate decomposition. In the practically relevant regime of moderate noise levels below $30\%$, posterior averaging has virtually no impact.
We further note in Fig.\ \ref{fig:benchmarks}, that dbtf features a similar or higher reconstruction accuracy with respect to the noisy training data. This indicates over-fitting and becomes particularly apparent for large noise levels and large tensor ranks. What distinguishes the decompositions of dbtf and TensOrM in this regime? 
In Fig.\ \ref{subfig:perturbation} we show the training performance under random perturbations of the inferred factors for a noise level of $40\%$ and a expected density of 10\%. Our method has a lower training accuracy but is more stable towards random perturbations of the parameters whereas the training accuracy of dbtf decreases rapidly. This shows that TensOrM converges to solutions of larger point-wise density in the parameter space that have better generalisation properties. Recently, this phenomenon has been studied in order to improve the generalisation of deep neural networks~\cite{Chaudhari2016}. Estimating posterior distributions, Bayesian techniques naturally assign more probability to such solutions.

\subsection{Model Selection}\label{subsec:model_selection}

A notorious challenge for latent variable models is the choice of dimensionality. Previously, \citet{Erdos2013a} have used the Minimum description length (MDL) principle for this task in Boolean Tensor Decomposition. We follow their derivation and compare to two approaches that our Bayesian treatment offers readily.
For the \textbf{Bayesian Occam's razor}, we start our inference procedure with a large latent dimensionality. After convergence of the Markov chain, we remove all latent dimensions that do not contribute to the likelihood. In these dimensions one of the factors is usually all zeros and the other factors are uniformly random. After removal we restart the burn-in procedure and repeat until only contributing dimensions remain. In the second approach, \textbf{cross validation}, we treat 20\% of the data as unobserved during training and choose the model dimensionality that achieves the highest posterior predictive accuracy on the held-out data. Results on random matrices for different noise levels are shown in Fig.\ref{fig:model_selection}, following the previously described simulation procedure. We find that both approaches clearly outperform MDL. In the case of noise-free observations the Bayesian Occam's razor features virtually perfect accuracy. For the more realistic scenario of moderate noise levels cross validation is superior.


\section{Real-world Applications}

Here we demonstrate the ability of TensOrM to infer meaningful, interpretable representations from real-world 3-way datasets of temporal interaction networks and temporal $\text{object}{\times}\text{property}$--relations.
With these moderately sized datasets of less than 100,000 data-points, sampling until convergence and drawing 50 samples takes only few seconds on a single core. Eventually, we turn to a large-scale biological example, analysing networks of relative gene-expression in cancer patients with more than 10 billion data points. Here, the inference procedure takes around 10 hours.

\begin{figure*}
	\centering
  \begin{minipage}[b]{0.5\textwidth}
	\subfigure[\textbf{Dynamic hospital interaction networks} -- Decomposition on the $13{\times}75{\times}75$ time-adjacency-tensor of interactions in a hospital ward. We only show one of the two equivalent individual-specific factors.]{
	\includegraphics[width=\linewidth]{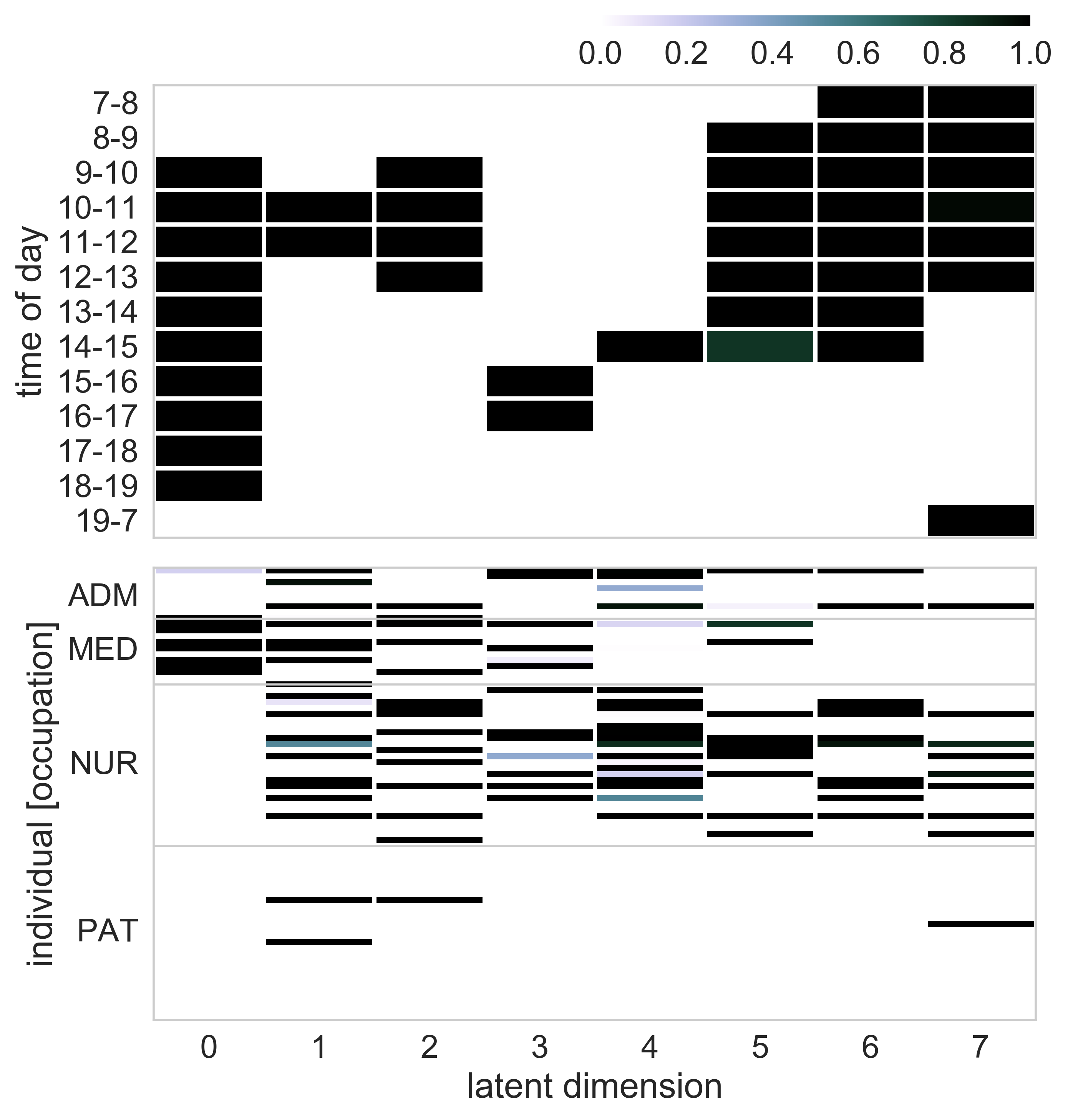}
	\label{subfig:hospital}}
  \end{minipage}\hspace{.03\linewidth}%
  \begin{minipage}[b]{0.45\textwidth}
	\subfigure[\textbf{Student dynamic seating throughout course} -- Decomposition of the $9{\times}7{\times}42$ week-seat-student tensor indicating seating positions for a 9-week university course on Android programming.]{
	\includegraphics[width=\linewidth]{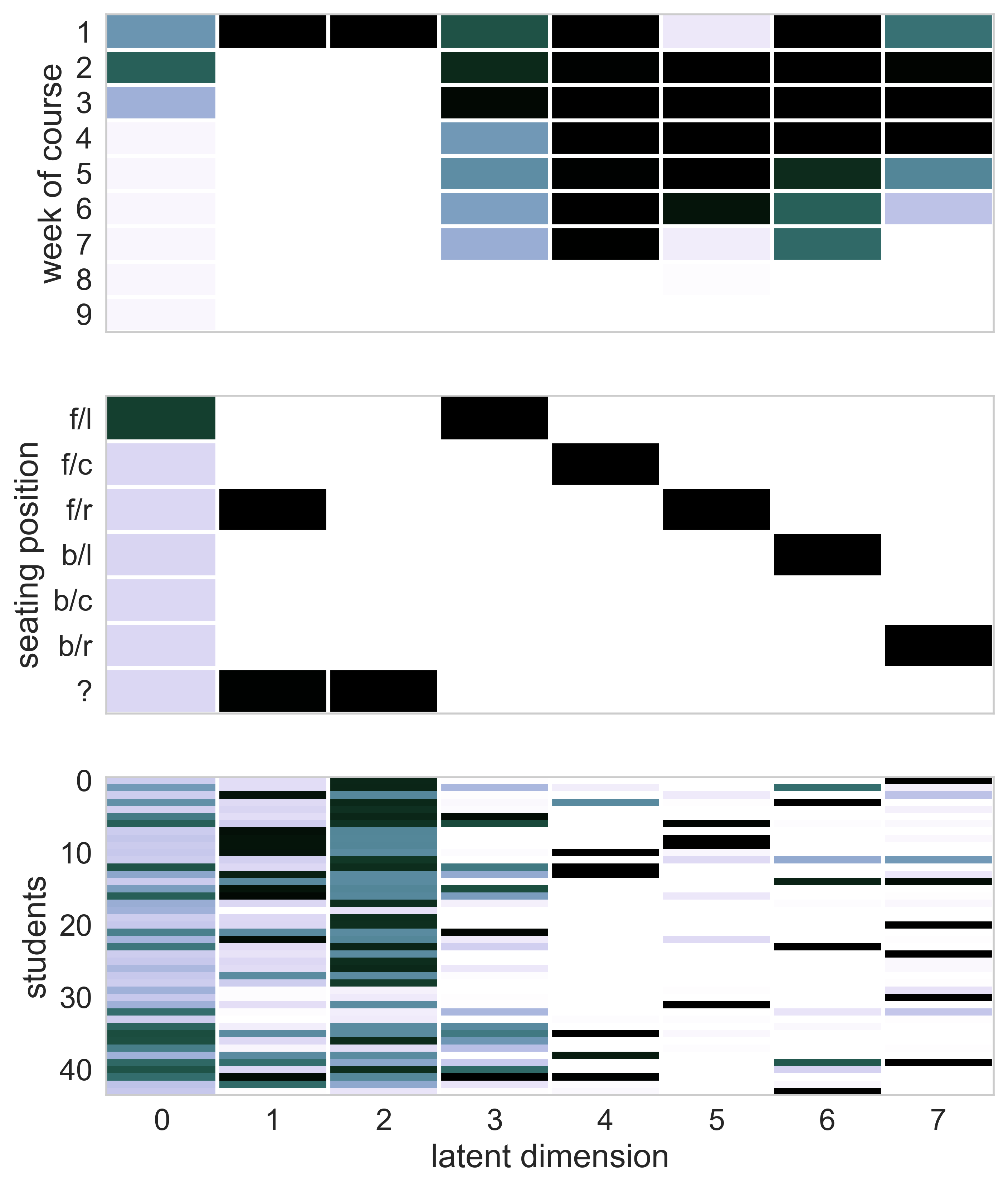}
	\label{subfig:students}}
  \end{minipage}
  \subfigure[\textbf{Representations of cancer patients} -- Each column corresponds to one out of approximately 8,000 cancer patients and indicates which of the latent properties of relative expression among approximately 2,000 genes they exhibit. Patients are ordered by type of cancer. See Figure 1 in the Supplementary Material for the corresponding representation from PCA on the continuous expression data, as well as for a legend of the disease types.]{
  \includegraphics[width=\linewidth]{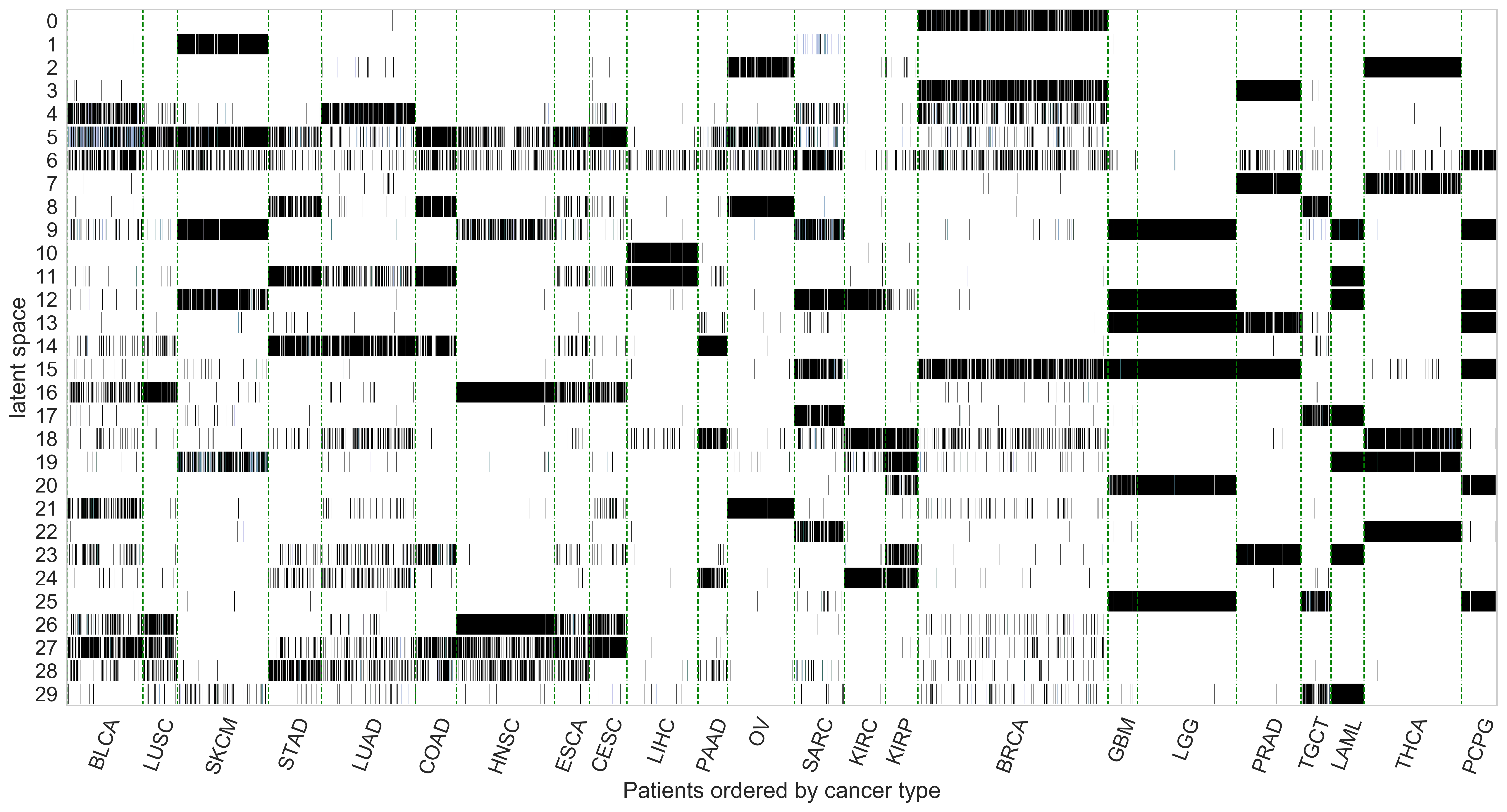}
  \label{subfig:cancer}
  }
	\caption{Real-world example of factors of TensOrMachine decomposition. Colours indicate posterior means.\label{fig:examples}}
\end{figure*}

\subsection{Hospital Ward Interaction Networks}\label{subsec:hospital}
Records of contact between pairs of individuals in a university hospital have originally been acquired to investigate transmission routes of infectious diseases~\cite{Vanhems2013}. Proximity sensor measurements were taken for 75 individuals in 20 second time windows across 5 days. In order to examine daily patterns, we group the time-points into 13 time-of-day windows. This leaves us with a binary $13{\times}75{\times}75$  time-adjacency tensor. We compute the decomposition, choosing 8 latent dimensions based on cross-validation as described in Section~\ref{sec:inference}. The predictive accuracy on the held-out test data is approximately $91\%$. Fig.\ \ref{subfig:hospital} shows posterior means of the time-specific and (one of the two equivalent) individual-specific factor matrices. Individuals are labelled as either administrative staff (ADM), medical doctors (MED), nurses (NUR) or patients (PAT).
Using this proximity data alone, we were able to recapitulate information about the work patterns of staff groups. For example, 10am-12pm and 3pm-5pm represent the main morning and afternoon clinic hours when all administrative, medical and nursing staff came into interaction, shown in latent dimensions 1 and 3. Whilst medical doctors are closely interacting throughout typical work hours, 9am-7pm, their network does not include any nurses or patients as found in dimension 0. This can be explained by frequent interaction in the doctor's room but also hints to a lack of interaction with the nursing staff, a phenomenon frequently discussed in the literature \citep[see e.g.][]{Manias2001}.
2pm-3pm indicates a period in which nurses and administrative staff are heavily interacting without participation of medical doctors, presumably indicating an afternoon break or daily joint meeting, see latent dimension 3. Nurses, however, were active throughout the day including being the main staffing body at night as dimension 7 shows.
In comparison, dbtf finds only 2 latent dimensions (0 and 5 in our analysis). 


\subsection{Student Seating Position}\label{subsec:students}
Our second example is part of the student-life dataset introduced by \citet{Harari2017} and given by records of the seating positions of students throughout a 9-week Android programming course. We partition the time-points into weeks of the course and the seating positions into six regions front/back - left/centre/right. An additional seating category is labelled with a question mark, where the seating coordinates were not provided but the students appeared in class. This yields a $9{\times}7{\times}42$ week-seat-student tensor. We choose seven latent dimensions, again, by optimising for the test-set likelihood with a reconstruction accuracy of approximately 92\% on the held out data. The decomposition is shown in Fig.\ \ref{subfig:students} and, once more, open to straightforward interpretation. The majority of students shows up to class in the first week of the course and sits scattered throughout the lecture hall, as can be seen in dimensions 0 and 1. The group of students that attends lectures most consistently, dimension 4, sits in the front-centre. The remaining groups describe subsets of students, each corresponding to exactly one of the four front/back left/right seating regions. The formation of some of these groups starts only in week 2 of the course and they all eventually stop to appear in class after 6-7 weeks. The input tensor confirms that only two students attend class in weeks 8 and 9.
Similar solutions are found by dbtf and shown in the SI (E) but lack the ability to characterise uncertainty around the inferred mode.

\subsection{Networks of Relative Gene Expression in Cancer}\label{subsec:cancer}

\newlength{\mytabskip}
\setlength{\mytabskip}{.125cm}
\begin{table}[t] %
\caption{Gene-set analysis of selected gene networks\label{tab:cancer} indicate their correspondence to cellular pathways. The sign indicates whether genes were relatively over/under expressed. 
}
\vskip 0.15in
\begin{center}
\begin{small}
\begin{sc}
\begin{tabular}{lll}
\toprule
\begin{tabular}{@{}l@{}}Dim.\end{tabular} & Cancer types & Pathways \\
\midrule
1 (-)  & SKCM                                                & Tight junctions \\ [\mytabskip]


4 (+)  & LUAD, BLCA                                          & Hippo$^*$ \\ [\mytabskip]

5 (+)  & \begin{tabular}{@{}l@{}} CESC, COAD, SKCM,\\ 
                          LUSC, ESCA, OV \end{tabular}       & Hippo, Wnt$^*$ \\ [\mytabskip]


7 (+)  & PRAD, THCA                                          & AGE-RAGE$^*$ \\ [\mytabskip]

9 (+)  & \begin{tabular}{@{}l@{}} LGG, PCPG, LAML,\\
                      GBM, SKCM, HNSC \end{tabular}          & Rap1 signalling \\ [\mytabskip]

11 (+) & \begin{tabular}{@{}l@{}}LAML, LIHC,\\
                                 COAD, STAD \end{tabular}    & \begin{tabular}{@{}l@{}}Actin\\ 
                                                                cytoskeleton\end{tabular} \\ [\mytabskip]


15 (+) & \begin{tabular}{@{}l@{}} PCPG, LGG, PRAD,\\ 
            GBM, BRCA, SARC \end{tabular}                    & \begin{tabular}{@{}l@{}}Focal\\
                                                               adhesion$^*$\end{tabular} \\ [\mytabskip]

16 (-) & \begin{tabular}{@{}l@{}}HNSC, LUSC,\\
                                 CESC, ESCA  \end{tabular}   & PI3K-Akt \\ [\mytabskip]

17 (+) & LAML, SARC, TGCT                                    & \begin{tabular}{@{}l@{}}Platelet\\
                                                                 activation\end{tabular} \\ [\mytabskip]    

18 (+) & \begin{tabular}{@{}l@{}}KIRC, KIRP,\\
                                 THCA, PAAD\end{tabular}     & Ras$^*$ \\ [\mytabskip]

24 (+) & KIRC, KIRP, PAAD                                    & Ras$^*$ \\ 
\bottomrule
\end{tabular}
\end{sc}
\end{small}
\end{center}
\vskip -0.1in
\end{table}

Gene expression profiling has been used extensively in cancer research, e.g.\ for the characterisation of cancer sub-types, for the stratification of patients or for the identification of therapeutic targets. 
Correlations in gene expression are at the basis of protein interaction in cellular pathways and latent variable models are frequently applied to extract biologically meaningful information from such data. 
We propose a novel way of analysing gene expression data that can be applied to any continuous measurement with $\text{object}{\times}\text{attribute}$ structure (here $\text{patient}{\times}\text{gene}$). The publicly available TCGA dataset \cite{Weinstein2013} contains gene expression measurements of a large variety of cancer patients across different types of cancer. After preprocessing, we are left with approximately 8,000 patients and 1,100 genes. We normalise the expression values for each gene across all patients to allow for comparison between genes. Then the data is transformed into a 3-way tensor using the relational encoding
\begin{align}
(\text{patient}, \text{gene}_i, \text{gene}_j) =
  \begin{cases}
    1;\text{ if expr}_i > \text{expr}_j \\
    0;\text{ if expr}_i < \text{expr}_j \\
    \text{unobserved; else}\;.
  \end{cases}
\end{align}
This provides an entirely new view of the data in terms of networks of relative expression.
For every latent dimension, the two gene specific factors indicate a subset of genes, where genes in the first subset are likely to be more expressed than genes in the second subset. The patient factor indicates which of these relational expression properties each patient exhibits. Importantly, the factors are amenable to distinct and intuitive interpretation with immediate biological relevance. We show the patient-specific factor in Fig.\ \ref{subfig:cancer} with patients ordered by cancer-type. We observe a remarkable disease specificity with many latent expression networks being exclusively present in virtually all cancers of certain types. In addition, some latent properties are scattered throughout cancers of various types, highlighting the heterogeneity of the disease. Application of a traditional method, principal component analysis (PCA) on the continuous $\text{patient}{\times}\text{gene}$ array, is shown in the Supplementary Material C. It lacks both, the interpretability and the degree of disease-type specificity. We use the representations of both methods as features for random forest classifiers. The predictive accuracy of the disease type is approximately 91\% for the binary TensOrM features, 86\% for continuous features from PCA and 81\% for binary dbtf features, shown in see SI (C, D).

The corresponding gene-specific factors characterise latent gene-networks but are more difficult to analyse for a non-specialist. In particular, the richness of the relational latent encoding can only partially be captured by traditional pathway analyses. Nevertheless, we investigate the biological plausibility of the inferred gene sets by running a gene set enrichment analysis of the genes in each factor dimension against the KEGG pathway database~\cite{Kanehisa2017}. Results are indicated in Tab.\ \ref{tab:cancer} and highlight the biological plausibility and significance of our analysis. We underline a few examples in the following. Firstly, Hippo signalling has been shown to be active in LUAD, LUSC, COAD, OV and LIHC~\cite{Harvey2013}, all recovered by Dims.\ 4, 5. Secondly, aberrant Wnt signalling is observed in many cancers but most prominently in COAD~\cite{Polakis2012} which is found in Dim.\ 5. Next, PRAD progression is known to be controlled by focal adhesion~\cite{Figel2011} as found in Dim.\ 15 as well as by the AGE-RAGE signalling pathway recovered in Dim.\ 7 \cite{Bao2015}. Finally, Ras signalling is aberrant in most tumours, but most significantly in PAAD \cite{Downward2003} as confirmed in dims.\ 18, 24.







\section{Conclusion}
We have introduced the first probabilistic approach to Boolean tensor decomposition. It reaches state-of-the-art performance and can readily deal with missing data which is ubiquitous in large datasets. Due to the particular combinatorial structure of the factor matrix posterior, sampling based inference scales to large datasets and enables the inference of posterior distributions over factor matrices, providing full uncertainty quantification. We further show that Boolean tensor decomposition leads to insightful latent representations and provides a novel view on the molecular basis of cancer by relationally encoding continuous expression data.

\section*{Acknowledgements}
This work was supported by The Alan Turing Institute under the
EPSRC grant EP/N510129/1. CY is supported by MRC grant MR/P02646X/1. TR is funded by EPSRC grant EP/G037280/1. TR thanks Taha Ceritli for helpful comments.


\bibliography{library}
\bibliographystyle{icml2018}


\onecolumn
\icmltitle{TensOrMachine -- Supplementary Information}

\icmlsetsymbol{equal}{*}
\begin{icmlauthorlist}
\icmlauthor{Tammo Rukat}{stats,ati}
\icmlauthor{Chris C.~Holmes}{stats}
\icmlauthor{Christopher Yau}{bmh}
\end{icmlauthorlist}


\icmlcorrespondingauthor{Tammo Rukat}{tammo.rukat@stats.ox.ac.uk}

\icmlkeywords{Matrix factorisation, Bayesian modelling, Genomics, Binary data}


\appendix
\section{Derivation of the conditionals}

Here we derive the full conditional distribution for a factor entry $f_{n_kl}$ as given in eq.\ (4) in the main paper. 
For constant priors, $p(f_{n_kl})=\text{const.}$, the conditional is given by normalising 
the likelihood for $f_{n_kl} \in \{0,1\}$. The likelihood has the form
\begin{equation}
	p(x_{[n]}|.) = \sigma\left[\lambda \tilde{x}_{[n]} \left( 1 - 2 
		\prod\limits_l \left[1-\prod\limits_{n \in [n]} f_{nl} \right] \right) \right]\label{eq:likelihood}
\end{equation}

and is factorial in the data-points $x_{[n]}$. Terms that do not depend on $f_{n_kl}$ cancel in the conditional and thus we take the product over all $[n]$ with $n_k$ fixed. Then normalising gives

\begin{align}
	p(f_{n_kl}=1|.) =
& \frac{\prod\limits_{\substack{[n],\;n_k \text{fixed}}} p(x_{[n]}|f_{n_kl}=1,\text{rest}) } 
	{\prod\limits_{\substack{[n],\;n_k \text{fixed}}} p(x_{[n]}|f_{n_kl}=1,\text{rest}) +
	 \prod\limits_{\substack{[n],\;n_k \text{fixed}}} p(x_{[n]}|f_{n_kl}=0,\text{rest})} \\
= & \sigma\left[ \sum\limits_{\substack{[n] \\ n_k \text{fixed}}} 
	\log \frac{p(x_{[n]}|f_{n_kl}=1,\text{rest})}{p(x_{[n]}|f_{n_kl}=0,\text{rest})}
	\right]\;. \label{eq:normalise}
\end{align}

Considering the term inside the logarithm in eq.\ \eqref{eq:normalise} and using eq.\ \eqref{eq:likelihood} we find
\begin{align}
\frac{p(x_{[n]}|f_{n_kl}=1,\text{rest})}{p(x_{[n]}|f_{n_kl}=0,\text{rest})} =
	\begin{cases}
		1;\;\text{if}\; \bigg(\prod\limits_{n \in [n]/n_k} f_{nl}\bigg)
		\prod\limits_{l'\neq l} \bigg( 1 - \prod\limits_{n \in [n]}f_{nl'}\bigg) = 0 \vspace{.2cm} \\ 
		\frac{1+\exp(\lambda \tilde{x}_{[n]})}{1+\exp(-\lambda \tilde{x}_{[n]})} = e^{\lambda \tilde{x}_{[n]}};\;
			\text{otherwise}\;.
	\end{cases}
\end{align}

The first equality describes all cases where the term inside the parenthesis in eq.\ \eqref{eq:likelihood} takes a value that
is independent of the value of $f_{n_kl}$. The second term follows in all other cases and by expanding the logistic sigmoid.
Hence, we can write eq.\ \eqref{eq:normalise} as
\begin{align}
		p(f_{n_kl}=1|.) = \sigma\left[ \lambda \sum\limits_{\substack{[n] \\ n_k \text{fixed}}} 
		\tilde{x}_{[n]} \overbrace{\bigg(\prod\limits_{n \in [n]/n_k} f_{nl}\bigg)
		\prod\limits_{l'\neq l} \bigg( 1 - \prod\limits_{n \in [n]}f_{nl'}\bigg)}^{M_{(n_k,l)\rightarrow [n]}}
	\right]\;.
\end{align}

\section{Cancer-type legend}
\begin{minipage}[c]{0.49\textwidth}
\begin{itemize}
\setlength{\itemsep}{1pt}
\setlength{\parskip}{0pt}
\setlength{\parsep}{0pt}
\item BLCA: Bladder Urothelial Carcinoma 
\item BRCA: Breast invasive carcinoma 
\item CESC: Cervical squamous cell carcinoma and endocervical adenocarcinoma 
\item COAD: Colon adenocarcinoma 
\item ESCA: Esophageal carcinoma 
\item GBM: Glioblastoma multiforme 
\item HNSC: Head and Neck squamous cell carcinoma 
\item KIRC: Kidney renal clear cell carcinoma 
\item KIRP: Kidney renal papillary cell carcinoma 
\item LAML: Acute Myeloid Leukemia
\item LGG: Brain Lower Grade Glioma
\end{itemize}
\end{minipage}
\begin{minipage}[c]{0.49\textwidth}
\begin{itemize}
\setlength{\itemsep}{1pt}
\setlength{\parskip}{0pt}
\setlength{\parsep}{0pt}
\item LIHC: Liver hepatocellular carcinoma 
\item LUAD: Lung adenocarcinoma 
\item  LUSC: Lung squamous cell carcinoma 
\item  OV: Ovarian serous cystadenocarcinoma 
\item  PAAD: Pancreatic adenocarcinoma 
\item  PCPG: Pheochromocytoma and Paraganglioma 
\item  PRAD: Prostate adenocarcinoma 
\item  SARC: Sarcoma 
\item  SKCM: Skin Cutaneous Melanoma 
\item  STAD: Stomach adenocarcinoma 
\item  TGCT: Testicular Germ Cell Tumours 
\item  THCA: Thyroid carcinoma
\end{itemize}
\end{minipage}

\section{Principal component analysis of gene expression data}

\begin{figure*}[h!]
\includegraphics[width=\linewidth]{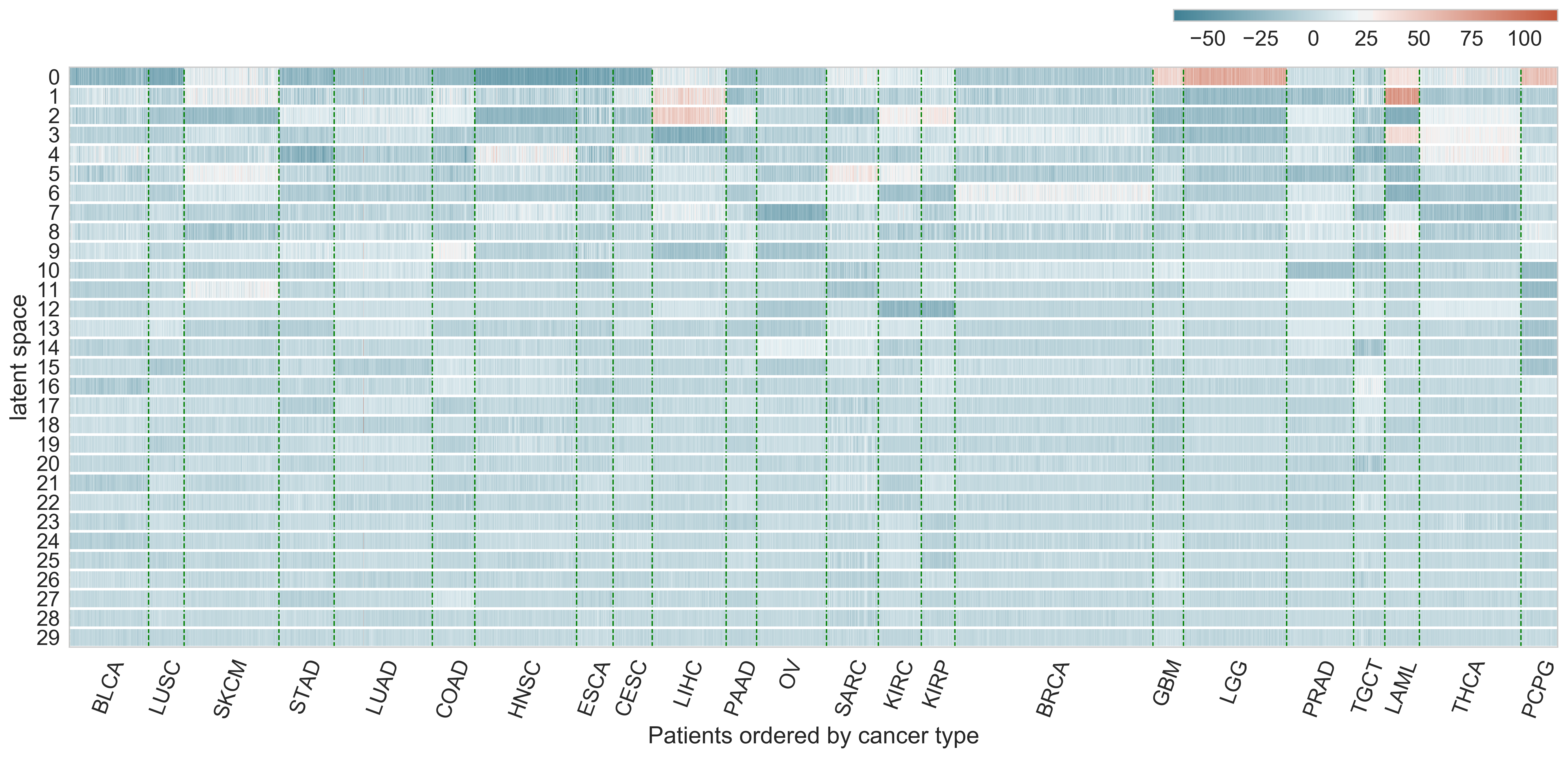}
\caption{Latent representation of the $\text{patient}{\times}\text{gene-expression}$ under principal component analysis and arranged in analogy to Fig.\ 4(c) in the main paper.}
\end{figure*}

\clearpage
\section{DBTF on Gene Expression Data}

\begin{figure*}[h!]
\includegraphics[width=\linewidth]{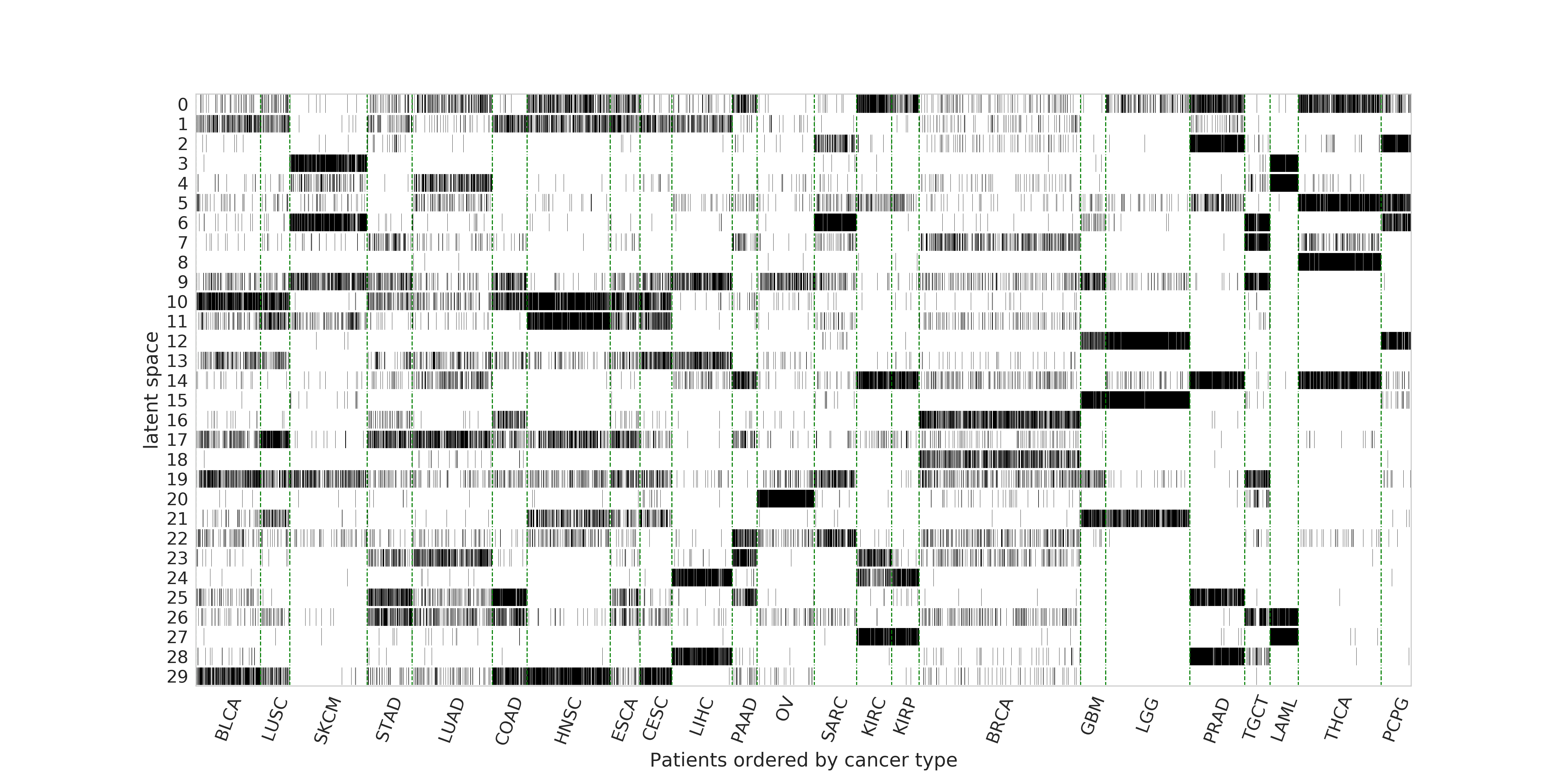}
\caption{Latent representation of the $\text{patient}{\times}\text{gene-expression}$ under dbtf arranged in analogy to Fig.\ 4(c) in the main paper.\label{fig:pancan_dbtf}}
\end{figure*}

Fig. \ref{fig:pancan_dbtf} shows patient representation for relative gene
expression networks computed by dbtf. The analysis is limited to 350 genes,
since our 32 core, 128 GB machine runs out of memory for larger analyses using
the implementation provided by Park et al. [2017]. Data that was treated as missing
in our original analysis is treated as 0 here. Comparing to Fig.\ 4(c) we observe
similar but noisier disease-specific patterns.

\clearpage
\section{Real World Data Analysed with dbtf}
\begin{figure*}[h!]
\includegraphics[width=.5\linewidth]{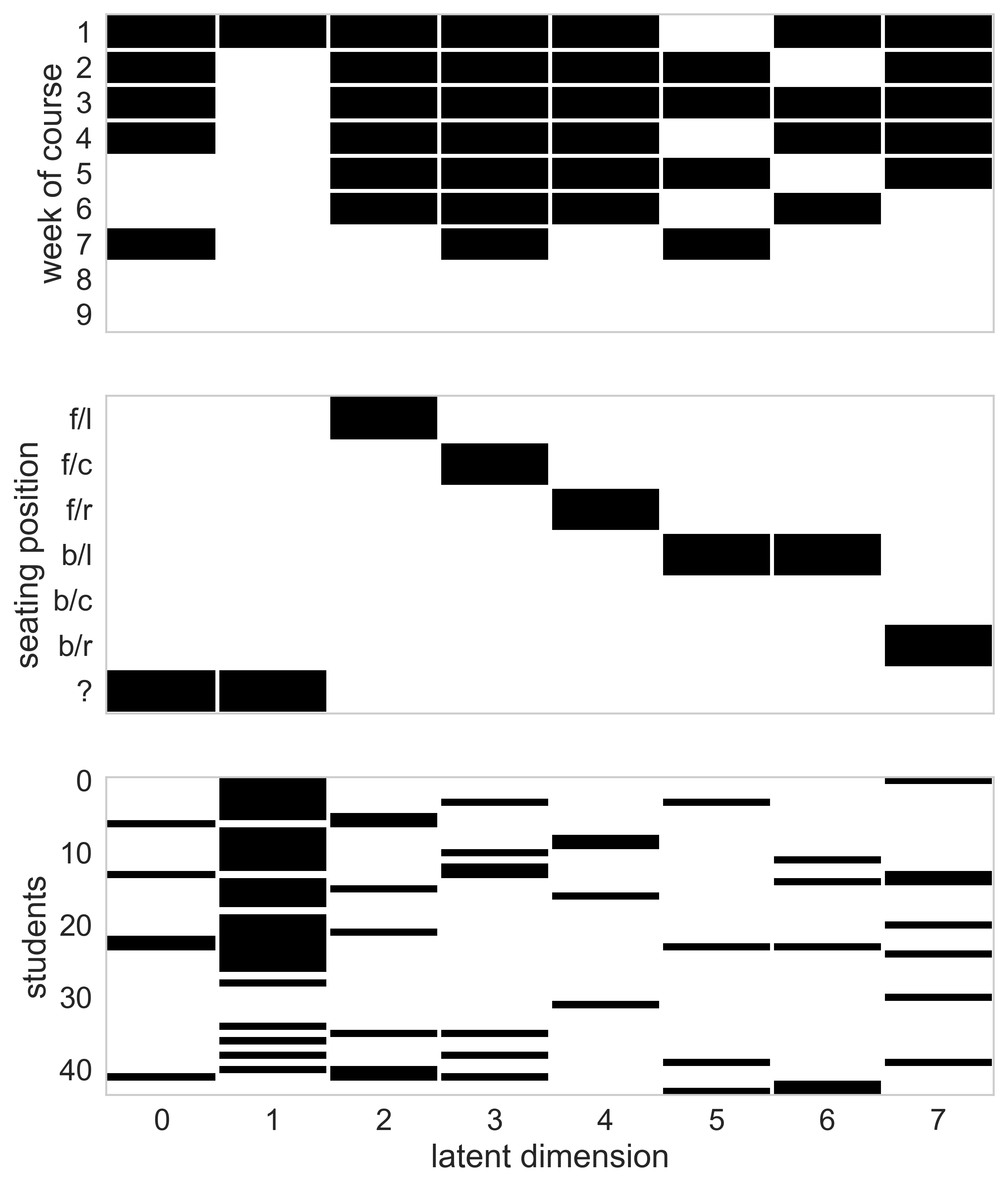}
\includegraphics[width=.5\linewidth]{students2.png}
\caption{Student seating throughout course, analysed with dbtf (left)
The latent representation are ordered to emphasis similarity to 
Fig.~4(b) in the main paper, a copy of which is shown on the right-hand-side for easier reference. The correspondence is as follows 
[dbtf latent dimensions $\rightarrow$ TensOrM latent dimensions]:
[0 $\rightarrow$ (0,1)],
[1 $\rightarrow$ 2],
[2 $\rightarrow$ 3], 
[3 $\rightarrow$ 4],
[4 $\rightarrow$ 5],
[(5,6), $\rightarrow$ 6],
[7 $\rightarrow$ 7].
\label{fig:student_dbtf}}
\end{figure*}

Fig. \ref{fig:student_dbtf} shows the results of dbtf, corresponding to
the analysis in Fig.~4(b) of the main paper. While this solution looks similar to a possible point estimate
of the TensOrM analysis, dbtf lack the ability to characterise posterior uncertainty
which is crucial in this example.
For the hospital data, dbtf infers only 3 latent dimensions. These correspond
to dimensions 2, 5, and 7 of the TensOrM analysis. 
Note, that we can not assess the predictive performance since dbtf is unable to 
deal with held-out data

\end{document}